\def\year{2018}\relax
\documentclass[letterpaper]{article} 
\usepackage{aaai18}  
\usepackage{times}  
\usepackage{helvet}  
\usepackage{courier}  
\usepackage{url}  
\usepackage{graphicx}  
\usepackage{amsmath}
\usepackage{algorithm}
\usepackage{algorithmic}
\usepackage{multirow}
\usepackage{tabularx}
\newcolumntype{Y}{>{\centering\arraybackslash}X}
\begin{document}
%
\title{Unsupervised Generative Adversarial Cross-modal Hashing}
\author{Jian Zhang, Yuxin Peng\thanks{Corresponding author.}, and Mingkuan Yuan\\
	Institute of Computer Science and Technology, Peking University\\
	Beijing 100871, China\\
	pengyuxin@pku.edu.cn\\
}
\maketitle
\begin{abstract}
Cross-modal hashing aims to map heterogeneous multimedia data into a common Hamming space, which can realize fast and flexible retrieval across different modalities. Unsupervised cross-modal hashing is more flexible and applicable than supervised methods, since no intensive labeling work is involved. However, existing unsupervised methods learn hashing functions by preserving inter and intra correlations, while ignoring the underlying manifold structure across different modalities, which is extremely helpful to capture meaningful nearest neighbors of different modalities for cross-modal retrieval. To address the above problem, in this paper we propose an \textit{Unsupervised Generative Adversarial Cross-modal Hashing approach (UGACH)}, which makes full use of GAN's ability for unsupervised representation learning to exploit the underlying manifold structure of cross-modal data. The main contributions can be summarized as follows: (1) We propose a \textbf{generative adversarial network to model cross-modal hashing} in an unsupervised fashion. In the proposed UGACH, given a data of one modality, the generative model tries to fit the distribution over the manifold structure, and select informative data of another modality to challenge the discriminative model. The discriminative model learns to distinguish the generated data and the true positive data sampled from correlation graph to achieve better retrieval accuracy. These two models are trained in an adversarial way to improve each other and promote hashing function learning. (2) We propose a \textbf{correlation graph} based approach to capture the underlying manifold structure across different modalities, so that data of different modalities but within the same manifold can have smaller Hamming distance and promote retrieval accuracy. Extensive experiments compared with 6 state-of-the-art methods on 2 widely-used datasets verify the effectiveness of our proposed approach.
\end{abstract}

\section{Introduction}
Multimedia retrieval has become an important application over the past decades, which can retrieve multimedia contents that users have interests in. However, it is a big challenge to retrieve multimedia data efficiently from large scale databases, due to the explosive growth of multimedia information. To address this issue, there are many hashing methods~\cite{imagehashsurvey,lsh_vldb,SSDH} proposed to accomplish efficient retrieval. The goal of hashing methods is to map high dimensional representations in the original space to short binary codes in the Hamming space. By using these binary hash codes, faster Hamming distance computation can be applied based on bit operations that can be implemented efficiently. Moreover, binary codes take much less storage compared with original high dimensional representations. 

There are large numbers of hashing methods applied to single modality retrieval~\cite{imagehashsurvey}, by which users can only retrieve data by a query with the same modality, such as text retrieval~\cite{textretrieval} and image retrieval~\cite{imagehashsurvey}. Nevertheless, single modality retrieval can not meet users' increasing demands, due to the different modalities of multimedia data. For example, by single modality retrieval, it is impracticable to search an image by using a textual sentence that describes the semantic content of the image. Therefore, cross-modal hashing has been proposed to meet this kind of retrieval demands in large scale cross-modal databases. Owing to the effectiveness and flexibility of cross-modal hashing, users can submit whatever they have to retrieve whatever they want ~\cite{overviewtcsvt,peng2017cross}.

``Heterogeneous gap'' is the key challenge of cross-modal hashing, which means the similarity of between different modalities cannot be measured directly. Consequently, some cross-modal hashing methods~\cite{cvh,pdh,cmfh,SCM,CMNNH} have been proposed to bridge this gap. Existing cross-modal hashing methods can be categorized into traditional methods and Deep Neural Networks (DNN) based methods. Moreover, traditional methods can be divided into unsupervised methods and supervised methods by whether semantic information is leveraged.

\textit{\textbf{Unsupervised cross-modal hashing methods}} usually project data from different modalities into a common Hamming space to maximize their correlations, which hold the similar idea with Canonical Correlation Analysis (CCA)~\cite{cca}. Song et al. propose Inter-Media Hashing (IMH)~\cite{imh} to establish a common Hamming space by preserving inter-media and intra-media consistency. Cross-view Hashing (CVH)~\cite{cvh} is proposed to consider both intra-view and inter-view similarities, which is an extension of image hashing method named Spectral Hashing (SH)~\cite{sh_nips}. Predictable Dual-view Hashing (PDH)~\cite{pdh} designs an objective function to keep the predictability of pre-generated binary codes. Ding et al. propose Collective Matrix Factorization Hashing (CMFH)~\cite{cmfh} to learn unified hash codes by collective matrix factorization. Composite Correlation Quantization (CCQ)~\cite{CCQ} jointly learns the correlation-maximal mappings that transform different modalities into an isomorphic latent space, and learns composite quantizers that convert the isomorphic latent features into compact binary codes.

\textit{\textbf{Supervised cross-modal hashing methods}} utilize labeled semantic information to learn hashing functions. Bronstein et al. propose Cross-Modality Similarity Sensitive Hashing (CMSSH)~\cite{CMSSH} to model hashing learning by a classification paradigm with a boosting manner. Wei et al. propose Heterogeneous Translated Hashing (HTH)~\cite{HTH}, which learns translators to align separate Hamming spaces of different modalities to perform cross-modal hashing. Semantic Correlation Maximization (SCM)~\cite{SCM} is proposed to learn hashing functions by constructing and preserving the semantic similarity matrix. Semantics-Preserving Hashing (SePH)~\cite{SePH} transforms the semantic matrix into a probability distribution and minimizes the KL-divergence in order to approximate the distribution with learned hash codes in Hamming space.  

\textit{\textbf{DNN based methods}} are inspired by the successful applications of deep learning, such as image classification~\cite{imagelcass}. Cross-Media Neural Network Hashing (CMNNH)~\cite{CMNNH} is proposed to learn cross-modal hashing functions by preserving intra-modal discriminative ability and inter-modal pairwise correlation. Cross Autoencoder Hashing (CAH)~\cite{CAH} is based on deep autoencoder structure to maximize the feature correlation and the semantic correlation between different modalities. Cao et al. propose Deep Visual-semantic Hashing (DVH)~\cite{DVH} as an end-to-end framework that combines both representation learning and hashing function learning. Jiang et al. propose Deep Cross-modal Hashing (DCMH)~\cite{DCMH}, which performs feature learning and hashing function learning simultaneously.

Compared with unsupervised paradigm, supervised methods use labeled semantic information that requires massive labor to collect, resulting in a high labor cost in real world applications. On the contrary, unsupervised cross-modal hashing methods can leverage unlabeled data to realize efficient cross-modal retrieval, which is more flexible and applicable in real world applications. 
However, most unsupervised methods learn hashing functions by preserving inter and intra correlations, while ignoring the underlying manifold structure across different modalities, which is extremely helpful to capture meaningful nearest neighbors of different modalities. To address this problem, in this paper, we exploit correlation information from underlying manifold structure of unlabeled data across different modalities to enhance cross-modal hashing learning.

Inspired by recent progress of Generative Adversarial Network (GAN)~\cite{gan,gancls,TUCH,irgan}, which has shown its ability to model the data distribution in an unsupervised fashion. In this paper, we propose an unsupervised generative adversarial cross-modal hashing (UGACH) approach. We design a graph-based unsupervised correlation method to capture the underlying manifold structure across different modalities, and a generative adversarial network to learn the manifold structure and further enhance the performance by an adversarial boosting paradigm. The main contributions of this paper can be summarized as follows:
\begin{itemize}
	\item We propose a \textbf{generative adversarial network to model cross-modal hashing} in an unsupervised fashion. In the proposed UGACH, given the data of any modality, the generative model tries to fit the distribution over the manifold structure, and selects informative data of another modality to challenge the discriminative model. While the discriminative model learns to distinguish the generated data and the true positive data sampled from correlation graph to achieve better retrieval accuracy. 
	\item We propose a \textbf{correlation graph} based learning approach to capture the underlying manifold structure across different modalities, so that data of different modalities but within the same manifold can have smaller Hamming distance and promote retrieval accuracy. We also integrate the proposed correlation graph into proposed generative adversarial network to provide manifold correlation guidance to promote the cross-modal retrieval accuracy.
\end{itemize}

Extensive experiments compared with 6 state-of-the-art methods on 2 widely-used datasets verify the effectiveness of our proposed approach.

The rest of this paper is organized as follows. In ``The Proposed Approach'' section, we present our UGACH approach in detail. The experimental results and analyses are reported in ``Experiment'' section. Finally, we conclude this paper in ``Conclusion'' section.

\begin{figure*}[!th]
	\centering
	\includegraphics[width=0.85\textwidth]{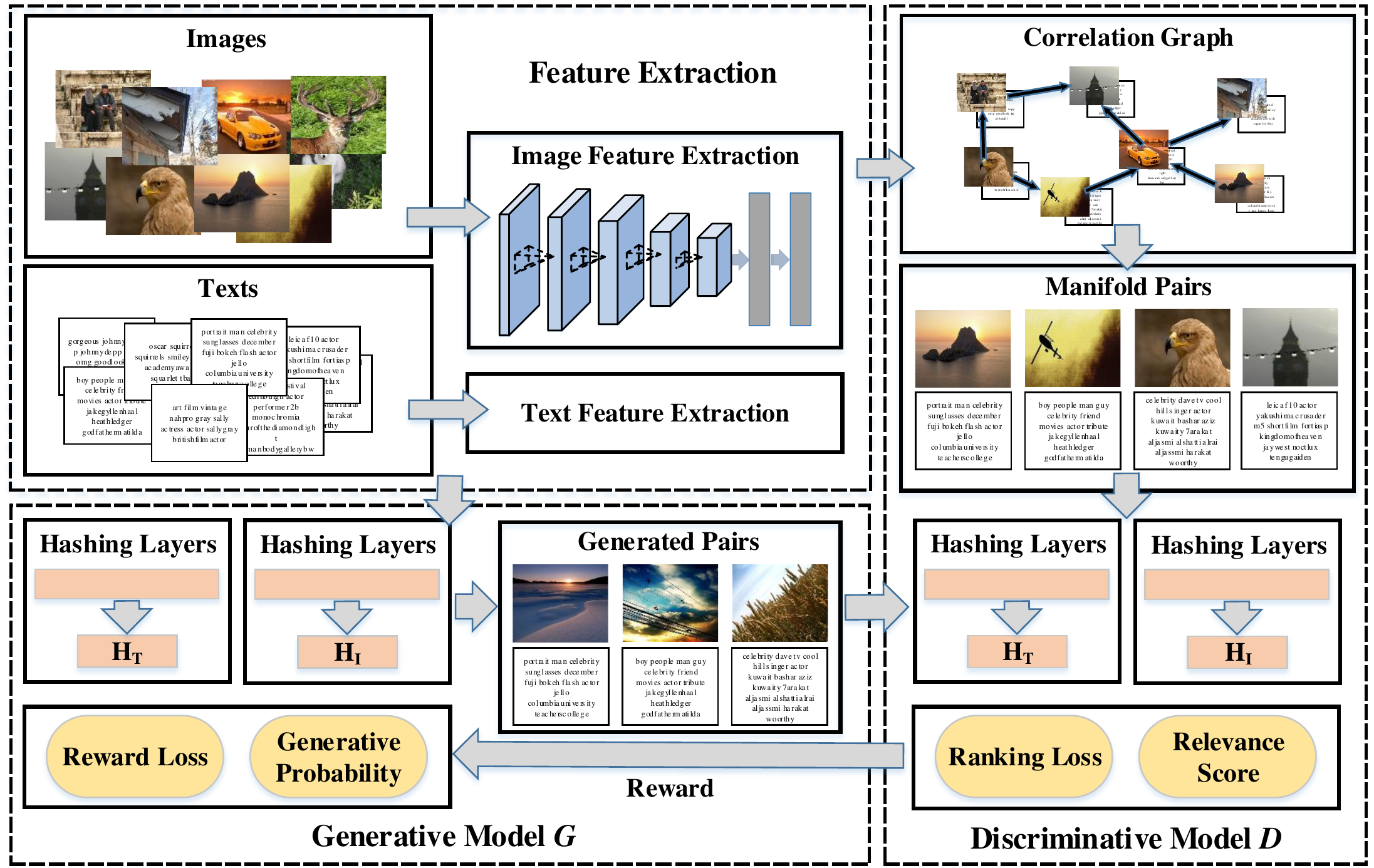}
	\caption{The overall framework of our proposed unsupervised generative adversarial cross-modal hashing approach (UGACH), which consists of feature extraction part, generative model $G$ and discriminative model $D$.}
	\label{framework}
\end{figure*}

\section{The Proposed Approach}
Figure~\ref{framework} presents the overview of our proposed approach, which consists of three parts, namely feature extraction, generative model $G$ and discriminative model $D$. The feature extraction part employs image feature and text feature extraction to represent unlabeled data of different modalities as original features. The detailed implementation of this part will be described at ``Experiment'' section. Given a data of one modality, $G$ attempts to select informative data from another modality to generate a pair of data and send them to $D$. In $D$, we construct a correlation graph, which can capture the manifold structure among the original features. $D$ receives the generated pairs as inputs, and also samples positive data from constructed graph to form a true manifold pair. Then $D$ tries to distinguish the manifold and generated pairs in order to get better discriminate ability. These two models play a minimax game to boost each other, and the finally trained $D$ can be used as cross-modal hashing model.

We denote the cross-modal dataset as $D=\{I,T\}$, where \textit{I} represents image modality and \textit{T} represents text modality. In this paper, $D=\{I,T\}$ is further split into a retrieval database $D_{db} = \{I_{db},T_{db}\}$ and a query set $D_{q}=\{I_{q},T_{q}\}$. In the retrieval database $D_{db}$, $I_{db} = \{i_p\}_{p=1}^n$, $T_{db} = \{t_p\}_{p=1}^n$ and $n$ is the number of data pairs. The retrieval database $D_{db}$ is also the training set. In the query set $D_{q}$, $I_{q} = \{i_p\}_{p=1}^t$, $T_{q} = \{t_p\}_{p=1}^t$ and $t$ is the number of data pairs. The aim is to learn cross-modal hashing functions to generate hash codes for data $C^I=H_I(I)$ and $C^T=H_T(T)$, so that different modal data that has similar semantic are close in the common Hamming space. In addition, we denote the hash code length as $l$. By the generated hash codes, we can retrieve the relevant data by a query of any modality from database of another modality efficiently.

\subsection{Generative Model}
The network of generative model has a two-pathway architecture, which receives the original features of both images and texts as inputs. Each pathway consists of a common representation layer and a hashing layer, whose implementations are two fully-connected layers. The first layer can convert the modality specific features to common representations, which make the instances of different modalities measurable in a common space. The representation produced by this layer can be denoted as follows:
\begin{equation}
\phi_c(x) = tanh(W_cx+b_c)
\end{equation}
where $x$ denotes the original features of images or texts, $W_c$ is the weight parameter of the common representation layer, and $b_c$ is the bias parameter.

The hashing layer can map the common representations into binary hash codes, so that the similarity between different modalities can be measured by fast Hamming distance calculation. The continuous real values of hash code is defined as:
\begin{equation}
h(x) = sigmoid(W_h\phi_c(x)+b_h)
\end{equation}
where $W_h$ is the weight parameter and $b_h$ is the bias parameter. Then we can get the binary codes by a thresholding function:
\begin{equation}
\label{binarycode}
b(x) = sgn(h_k(x)-0.5), \quad k=1,2,\cdots,l
\end{equation}
where $l$ denotes the hash code length. Considering that it is hard to optimize binary codes directly, we use relaxed continuous real valued hash codes $h(x)$ in the training process.

Given a data of one modality, the goal of generative model $G$ is to fit the distribution over the manifold structure and select informative data of another modality to challenge the discriminative model. The generative probability of $G$ is $p_\theta(x^U|q)$, which is the foundation to select relevant instance of one modality from unpaired data when given a query of another modality. For example, given a image query $q_i$, the generative model tries to select relevant text $t^U$ from $T_{db}$. 

The generative probability $p_{\theta}(x^U|q)$ is defined as a softmax function:
\begin{equation}
	\label{generativedef}
	p_{\theta}(x^U|q) = \frac{\exp(-\|h(q)-h(x^U)\|^2)}{\sum_{x^U}\exp(-\|h(q)-h(x^U)\|^2)}
\end{equation}
Given a query, we can use equation~(\ref{generativedef}) to calculate the probability of each candidate, which indicates the possibility of becoming a relevant sample.

\subsection{Discriminative Model}

The network structure of discriminative model is same as the generative model. The input of this network is the generated pairs by generative model, and the manifold pairs provided by a correlation graph. The goal of discriminative model is to distinguish whether an input pair is generated or from the correlation graph. 

First of all, we introduce the correlation graph in the discriminative model. We propose a correlation graph to guide the training of discriminative model. The correlation graph can capture the underlying manifold structure across different modalities, so that data of different modalities but within the same manifold can have small Hamming distance and promote retrieval accuracy.

Specifically, we first construct undirected graphs $Graph_i=(V,W_i)$ and $Graph_t=(V,W_t)$ for image and text modality respectively, where $V$ denotes the vertices and $W_i$ and $W_t$ are the similarity matrix. $W_i$ is defined as follows:
\begin{equation}
\label{equknn}
w(p,q)=\left\{ \begin{array}{rrr}
1 :&  x_p \in NN_k(x_q)\\
0 :&  otherwise \\
\end{array}
\right.
\end{equation}
where $NN_k(x_q)$ denotes the $k$-nearest neighbors of $x_q$ in the training set, similarly we can define $W_t$. Then we sample the data of the true distributions based on the constructed graph. For real pair $p_{true}(x|q^j)$ provided by the dataset, we select $x_k$ as the true relevant instance of given query $q^j$ and associate them as manifold pair $p_{manifold}(x_k|q^j)$, when $w(k,j)=1$. It is noted that pairwise information naturally exists in cross-modal data, thus if the corresponding text $t_k$ is within the same manifold with text query $q^j$, the paired image $i_k$ is also in the same manifold with $q^j$ and vice versa. By this definition, we intend to utilize the underlying data manifold of different modality to guide the training of discriminative model. Intuitively, we want the data of different modality but within the same manifold to have small Hamming distance (e.g. small Hamming distance between text query $q^j$ and image $i_k$).

After receiving the generated and manifold pairs, discriminative model predicts a relevance score between each pair as the judgment result. So the relevance score between instance $x$ and its query $q$ is defined as $f_\phi(x,q)$. The goal of discriminative model is to distinguish the true relevant data (manifold pairs) and non-relevant data (generated pairs) for a query $q$ accurately.

The relevance score of $f_\phi(x^G,q)$ is defined by triplet ranking loss as follows:
\begin{equation}
\label{triplet_u}
\begin{split}
f_\phi(x^G,q) = \max(0, m&+\|h(q)-h(x^M)\|^2\\
&-\|h(q)-h(x^G)\|^2)
\end{split}
\end{equation}
where $x^M$ is a manifold paired instance with query $q$ selected from the correlation graph, $x^G$ is the selected instance by generative model, and $m$ is a margin parameter which is set to be $1$ in our proposed approach. The above equation means that we want the distance between manifold pair $(q,x^M)$ smaller than that of generated pair $(q,x^G)$ by a margin $m$, so that the discriminative model can draw a clear distinguishing line between the manifold and generated pairs.

Then discriminative model $D$ uses the relevance score to produce predicted probability of instance $x$ by a sigmoid function:
\begin{equation}
	\label{discrminativedef}
	\begin{split}
		D(x|q) &= sigmoid(f_\phi(x,q)) = \frac{\exp(f_\phi(x,q))}{1+\exp(f_\phi(x,q))}
	\end{split}
\end{equation}

The generative model tries to select informative data to challenge the discriminative model, which limits its capability to perform cross-modal retrieval. By contrast, the discriminative model is suitable for retrieving data across different modalities, after being promoted greatly by the generative model. Therefore after the proposed UGACH is trained, we use the discriminative model to perform cross-modal retrieval via produced hash codes.

\subsection{Adversarial Learning}
Given the definitions of the generative and discriminative models, we can conduct a minimax game for training them. Given a query, the generative model attempts to generate a pair which is close to the manifold pair to fool the discriminative model. The discriminative model tries to distinguish between manifold pair sampled from the correlation graph and the generated pair, which forms an adversarial process against the generative model. Inspired by the GAN~\cite{gan}, this adversarial process can be defined:
\begin{equation}
\label{minimaxgame_ti}
\begin{split}
\mathcal{V}(G,D) = &\min_{\theta}\max_{\phi}\sum_{j=1}^{n}(E_{x\sim p_{true}(x^M|q^j)}[log(D(x^M|q^j))]\\
&+E_{x\sim p_{\theta}(x^G|q^j)}[log(1-D(x^G|q^j))])
\end{split}
\end{equation}

The generative and discriminative models can be learned iteratively by maximizing and minimizing the above object function. As general training process, the discriminative model tries to \textit{maximize} equation~(\ref{minimaxgame_ti}), while the generative model attempts to \textit{minimize} equation~(\ref{minimaxgame_ti}) and fit the distribution over the manifold structure. The learning process of the discriminative model is fixed when the generative model can be trained as follows:
\begin{scriptsize}
	\begin{equation}
	\label{gen_opt}
	\begin{split}
	\theta^\ast=&\arg\min_{\theta}\sum_{j=1}^{n}(E_{x\sim p_{true}(x^G|q^j)}[log(sigmoid(f_{\phi^\ast}(x^M,q^j)))]\\
	&+E_{x\sim p_{\theta}(x^G|q^j)}[log(1-sigmoid(f_{\phi^\ast}(x^G,q^j)))])
	\end{split}
	\end{equation}
\end{scriptsize}
where $f_{\phi^\ast}$ denotes the discriminative model at previous iteration. The traditional GAN uses continuous noise vector to generate new data and is trained by stochastic gradient descent algorithm. By contrast, the generative model of our proposed UGACH selects data from unlabeled data to generate pairs and can not be optimized continuously due to the discrete selective strategy. We utilize reinforcement learning based parameters update policy to train the generative model as follows:
\begin{small}
	\begin{equation}
	\label{gen_opt_reinforce}
	\begin{split}
	&\nabla_\theta E_{x\sim p_{\theta}(x^G|q^j)}[log(1+\exp(f_\phi(x^G,q^j)))]\\
	&=\sum_{k=1}^{m}\nabla_\theta p_{\theta}(x_k^G|q^j)log(1+\exp(f_\phi(x_k^G,q^j)))\\
	&=\sum_{k=1}^{m}p_{\theta}(x_k^U|q^j)\nabla_\theta logp_{\theta}(x_k^G|q^j)log(1+\exp(f_\phi(x_k^G,q^j)))\\
	&=E_{x\sim p_{\theta}(x^G|q^j)}[\nabla_\theta logp_{\theta}(x^G|q^j)log(1+\exp(f_\phi(x^G,q^j)))]\\
	&\simeq \frac{1}{m}\sum_{k=1}^{m}\nabla_\theta logp_{\theta}(x_k^G|q^j)log(1+\exp(f_\phi(x_k^G,q^j)))
	\end{split}
	\end{equation}
\end{small}
where $k$ denotes the $k$-th instance selected by generative model according to a query $q^j$. From the perspective of reinforcement learning, according to the environment $q^k$, $x_k^G$ is the action taken by policy $logp_{\theta}(x_k^G|q^j)$, and $log(1+\exp(f_\phi(x_k^G,q^j)))$ acts as the reward, which encourages the generative model to select data close to the distribution over manifold structure. Finally the trained discriminative model can be used to generate binary codes for any input data of any modality, and cross modal retrieval can be performed by fast Hamming distance computation between query and each data in the database.

\section{Experiments}
In this section, we present the experimental results of our proposed UGACH approach. We first introduce the datasets, evaluation metrics and implementation details. Then we compare and analyze the results of UGACH with 6 state-of-the-art methods and 2 baseline methods.
\subsection{Dataset}
In the experiments, we conduct cross-modal hashing on 2 widely-used datasets: NUS-WIDE~\cite{nuswide} and MIRFLICKR~\cite{mirflickr}.
\begin{itemize}
	\item \textbf{NUS-WIDE} dataset~\cite{nuswide} is a relatively large-scale image/tag dataset with 269498 images. Each image has corresponding textual tags, which are regarded as the text modality in our experiments. NUS-WIDE dataset has 81 categories, but there are overlaps among the categories. Following~\cite{SePH}, we select the 10 largest categories and the corresponding 186557 images. We take $1\%$ data of NUS-WIDE dataset as the query set, and the rest as the retrieval database. We randomly selected 5000 images as training set for the supervised methods. We represent each image by 4096 dimensional deep features extracted from 19-layer VGGNet, and each text by 1000 dimensional BoW.
	\item \textbf{MIRFlickr} dataset~\cite{mirflickr} has 25000 images collected from Flickr, which has 24 categories. Each image is also associated with text tags. Following~\cite{SePH}, we take $5\%$ of the dataset as the query set and the remaining as the retrieval database. We also randomly select 5000 images as training set for supervised methods. Similarly, we represent each image by 4096 dimensional deep features extracted from 19-layer VGGNet, and each texts by 1000 dimensional BoW.
\end{itemize}

\subsection{Compared Methods}
In order to verify the effectiveness of our proposed approach, there are 4 unsupervised methods and 2 supervised methods compared in the experiment, including unsupervised methods CVH~\cite{cvh}, PDH~\cite{pdh}, CMFH~\cite{cmfh} and CCQ~\cite{CCQ}, and supervised methods CMSSH~\cite{CMSSH} and SCM~\cite{SCM}. Besides state-of-the-art methods, we also compare our UGACH approach with 2 baseline methods to verify the effectiveness of our contributions. 
\begin{itemize}
	\item \textit{Baseline}: We design a baseline method without the correlation graph and adversarial training, we denote this method as $Baseline$. It is implemented by training the discriminative model alone with a triplet ranking loss in equation~(\ref{triplet_u}), where the positive data is only the paired data provided by cross-modal datasets.
	\item \textit{Baseline-GAN}: We add the adversarial training to \textit{Baseline}, which means that we further promote discriminative model in \textit{Baseline} by adversarial training defined in equation~(\ref{minimaxgame_ti}).
\end{itemize}
Comparing \textit{Baseline-GAN} with \textit{Baseline}, we can verify the effectiveness of our proposed generative adversarial network for cross-modal hashing, and comparing our final approach UGACH with \textit{Baseline-GAN}, we can verify the effectiveness of proposed correlation graph.

\begin{table*}[tbh]
	\centering
	\caption{The MAP scores of two retrieval tasks on NUS-WIDE dataset with different lengths of hash codes.}
	\label{nusmap}
	\begin{tabularx}{0.85\textwidth}{c|Y|Y|Y|Y|Y|Y|Y|Y}
		\hline
		\multirow{2}{*}{Methods} & \multicolumn{4}{c|}{image$\rightarrow$text}      & \multicolumn{4}{c}{text$\rightarrow$image}      \\ \cline{2-9} 
		& 16    & 32    & 64    & 128   & 16    & 32    & 64    & 128   \\ \hline
		CVH~\cite{cvh}          & 0.458 & 0.432 & 0.410 & 0.392 & 0.474 & 0.445 & 0.419 & 0.398 \\ 
		PDH~\cite{pdh}          & 0.475 & 0.484 & 0.480 & 0.490 & 0.489 & 0.512 & 0.507 & 0.517 \\ 
		CMFH~\cite{cmfh}        & 0.517 & 0.550 & 0.547 & 0.520 & 0.439 & 0.416 & 0.377 & 0.349 \\ 
		CCQ~\cite{CCQ}          & 0.504 & 0.505 & 0.506 & 0.505 & 0.499 & 0.496 & 0.492 & 0.488 \\ \hline
		CMSSH~\cite{CMSSH}      & 0.512 & 0.470 & 0.479 & 0.466 & 0.519 & 0.498 & 0.456 & 0.488 \\ 
		SCM\_orth~\cite{SCM}          & 0.389 & 0.376 & 0.368 & 0.360 & 0.388 & 0.372 & 0.360 & 0.353 \\ 
		SCM\_seq~\cite{SCM}           & 0.517 & 0.514 & 0.518 & 0.518 & 0.518 & 0.510 & 0.517 & 0.518 \\ \hline
		Baseline                 & 0.540 & 0.537 & 0.573 & 0.598 & 0.554 & 0.555 & 0.583 & 0.608 \\ 
		Baseline-GAN             & 0.575 & 0.594 & 0.602 & 0.623 & 0.580 & 0.609 & 0.617 & 0.629 \\ 
		\textbf{UGACH (Ours)}          & \textbf{0.613} & \textbf{0.623} & \textbf{0.628} & \textbf{0.631} & \textbf{0.603} & \textbf{0.614} & \textbf{0.640} & \textbf{0.641} \\ \hline
	\end{tabularx}
\end{table*}

\begin{table*}[tbh]
	\centering
	\caption{The MAP scores of two retrieval tasks on MIRFlickr dataset with different lengths of hash codes.}
	\label{mirmap}
	\begin{tabularx}{0.85\textwidth}{c|Y|Y|Y|Y|Y|Y|Y|Y}
		\hline
		\multirow{2}{*}{Methods} & \multicolumn{4}{c|}{image$\rightarrow$text}      & \multicolumn{4}{c}{text$\rightarrow$image}      \\ \cline{2-9} 
		& 16    & 32    & 64    & 128   & 16    & 32    & 64    & 128   \\ \hline
		CVH~\cite{cvh}           & 0.602 & 0.587 & 0.578 & 0.572 & 0.607 & 0.591 & 0.581 & 0.574 \\ 
		PDH~\cite{pdh}           & 0.623 & 0.624 & 0.621 & 0.626 & 0.627 & 0.628 & 0.628 & 0.629 \\ 
		CMFH~\cite{cmfh}         & 0.659 & 0.660 & 0.663 & 0.653 & 0.611 & 0.606 & 0.575 & 0.563 \\ 
		CCQ~\cite{CCQ}           & 0.637 & 0.639 & 0.639 & 0.638 & 0.628 & 0.628 & 0.622 & 0.618 \\ \hline
		CMSSH~\cite{CMSSH}       & 0.611 & 0.602 & 0.599 & 0.591 & 0.612 & 0.604 & 0.592 & 0.585 \\ 
		SCM\_orth~\cite{SCM}          & 0.585 & 0.576 & 0.570 & 0.566 & 0.585 & 0.584 & 0.574 & 0.568 \\ 
		SCM\_seq~\cite{SCM}           & 0.636 & 0.640 & 0.641 & 0.643 & 0.661 & 0.664 & 0.668 & 0.670 \\ \hline
		Baseline                    & 0.619 & 0.631 & 0.633 & 0.646 & 0.625 & 0.635 & 0.634 & 0.649 \\  
		Baseline-GAN                & 0.630 & 0.643 & 0.651 & 0.664 & 0.660 & 0.657 & 0.670 & 0.688 \\ 
		\textbf{UGACH (Ours)}     & \textbf{0.685} & \textbf{0.693} & \textbf{0.704} & \textbf{0.702} & \textbf{0.673} & \textbf{0.676} & \textbf{0.686} & \textbf{0.690}  \\ \hline
	\end{tabularx}
\end{table*}

\subsection{Retrieval Tasks and Evaluation Metrics}
In the experiments, two retrieval tasks are performed: retrieving text by image query (image$\rightarrow$text) and retrieving images by text query (text$\rightarrow$image). Specifically, we first obtain the hash codes for the images and texts in the query and retrieval database with our UGACH approach and all the compared methods. Then we take one of the images as query, compute the Hamming distance with all text in retrieval database, and evaluate the ranking list by 3 evaluation metrics to measure the retrieval effectiveness: Mean Average Precision (MAP), precision recall curve (PR-curve) and precision at top $k$ returned results (top$K$-precision), which are defined as follows:
\begin{itemize}
	\item The MAP scores are computed as the mean of average precision (AP) for all queries, and AP is computed as:
	\begin{equation}
	AP=\frac{1}{R}\sum_{k=1}^{n}\frac{k}{R_k}\times rel_k
	\end{equation}
	where $n$ is the size of database, $R$ is the number of relevant images in database, $R_k$ is the number of relevant images in the top $k$ returns, and $rel_k=1$ if the image ranked at $k$-th position is relevant and 0 otherwise. 
	\item Precision recall curve (PR-curve): The precision at certain level of recall of the retrieved ranking list, which is widely used to measure the information retrieval performance.
	\item Precision at top $k$ returned results (top$K$-precision): The precision with respect to different numbers of retrieved samples from the ranking list.
\end{itemize}
It should be noted that the MAP score is computed for all the retrieval results with 4 different lengths of hash codes, while PR-curve and top$K$-precision are evaluated on 128 bit hash codes.

\subsection{Implementation Details}
In this section, we present the implementation details of our UGACH in the experiments. We take 4096 dimensional feature extracted from 19-layer VGGNet for images, and use the 1000 dimensional BoW feature for texts. We implement the proposed UGACH by tensorflow\footnote{https://www.tensorflow.org}. The dimension of common representation layer is set to be 4096, while the hashing layer's dimension is set to be the same as hash code length.

Moreover, we train the proposed UGACH in a mini-batch way and set the batch size as 64 for discriminative and generative models. We train the proposed UGACH iteratively. After the discriminative model is trained in 1 epoch, the generative model respectively will be trained in 1 epoch. The learning rate of UGACH is decreased by a factor of 10 each two epochs, while it is initialized as $0.01$.

For the compared methods, we apply the implementations provided by their authors, and follow their best settings to preform the experiments. And it is noted that for a fair comparison between different methods, we use the same image and text features for all compared methods.

\begin{figure*}[tbh]
	\centering
	\includegraphics[width=0.8\textwidth]{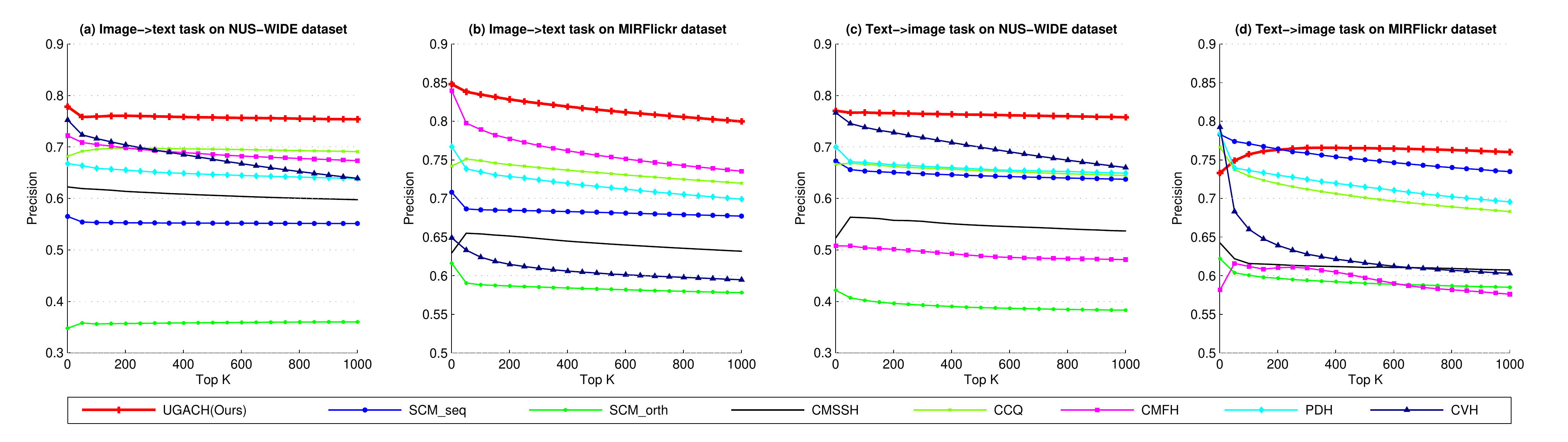}
	\caption{The top$K$-precision curves with 128 bit hash codes. The left two figures present the result of image$\rightarrow$text task on NUS-WIDE and MIRFlickr datasets, while the right two figures show the result of text$\rightarrow$image task.} 
	\label{topk}
\end{figure*}

\begin{figure*}[tbh]
	\centering
	\includegraphics[width=0.8\textwidth]{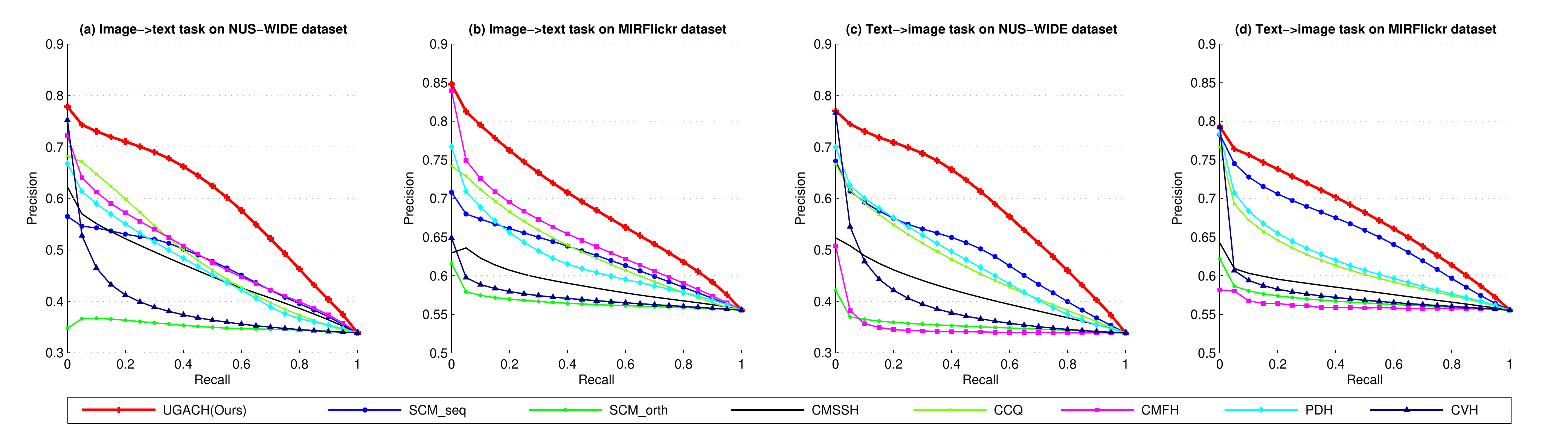}
	\caption{The precision-recall curves with 128 bit hash codes. The left two figures present the result of image$\rightarrow$text task on NUS-WIDE and MIRFlickr datasets, while the right two figures show the result of text$\rightarrow$image task.} 
	\label{prcurve}
\end{figure*}

\subsection{Experiment Results}
Tables~\ref{nusmap} and~\ref{mirmap} show the MAP scores of our UGACH and the compared methods on NUS-WIDE and MIRFlickr datasets. 

Compared with state-of-the-art methods, it can be seen that our proposed UGACH approach achieves the best retrieval accuracy on all 2 datasets. For convenience, we categorize these result tables into three parts: unsupervised compared methods, supervised compared methods and baseline methods. On NUS-WIDE dataset, our proposed UGACH keeps the best average MAP score of $0.624$ on image$\rightarrow$text and $0.625$ on text$\rightarrow$image tasks. Compared with the best unsupervised methods CCQ~\cite{CCQ}, our UGACH achieves an inspiring accuracy improvement from $0.505$ to $0.624$ on image$\rightarrow$text task, and improves the average MAP score from $0.494$ to $0.625$ on text$\rightarrow$image task. Even compared with supervised methods SCM\_seq~\cite{SCM}, our UGACH also improves average MAP scores from $0.517$ to $0.624$ on image$\rightarrow$text task, and from $0.516$ to $0.625$ on text$\rightarrow$image task. We can observe the similar trends on MIRFlickr dataset from Tables~\ref{mirmap}.

Figures~\ref{topk} and~\ref{prcurve} show the top$K$-precision and precision-recall curves on the two datasets with 128 bit code length. We can observe that on both image$\rightarrow$text and text$\rightarrow$image tasks, UGACH achieves the best accuracy among all compared unsupervised methods. And UGACH even achieves better retrieval accuracy than compared supervised methods on most of the evaluation metrics, which further demonstrates the effectiveness of our proposed approach.
 
Compared with 2 baseline methods on NUS-WIDE dataset, we can observe that \textit{Baseline-GAN} has an improvement of $0.037$ and $0.034$ on two tasks, which demonstrates the effectiveness of our proposed generative adversarial network for cross-modal hashing. Compared our proposed UGACH with \textit{Baseline-GAN}, we can observe an improvement of $0.025$ and $0.016$ on two tasks, which demonstrates the effectiveness of our proposed correlation graph. Similar trends can be also observed on MIRFlickr dataset.

\section{Conclusion}
In this paper, we have proposed an Unsupervised Generative Adversarial Cross-modal Hashing approach (UGACH), which intends to make full use of GAN's ability of unsupervised representation learning to exploit the underlying manifold structure of cross-modal data.
On one hand, we propose a generative adversarial network to model cross-modal hashing in an unsupervised fashion. In the proposed UGACH, the generative model tries to fit the distribution over the manifold structure, and select informative data of another modality to challenge the discriminative model. While the discriminative model learns to preserve traditional inter correlation, and the manifold correlations provided by generative model to achieve better retrieval accuracy. Those two models are trained in an adversarial way to improve each other and achieve better retrieval accuracy.
On the other hand, we propose a graph based correlation learning approach to capture the underlying manifold structure across different modalities, so that data of different modalities but within the same manifold can have smaller Hamming distance and promote retrieval accuracy. Experiments compared with 6 state-of-the-art methods on 2 widely-used datasets verify the effectiveness of our proposed approach.

The future works lie in two aspects. Firstly, we will focus on extending our approach to support retrieval across multiple modalities, such as cross-modal retrieval across image, text, video and audio. Secondly, we attempt to extend current framework to other scenarios such as image caption to verify its versatility.

\section{Acknowledgments}
This work was supported by National Natural Science Foundation of China under Grant 61771025 and Grant 61532005.

\bibliographystyle{aaai}
\bibliography{AAAI_UCHGAN}

\end{document}